\documentclass[letterpaper]{article} 
\usepackage{aaai25}  
\usepackage{times}  
\usepackage{helvet}  
\usepackage{courier}  
\usepackage[hyphens]{url}  
\usepackage{graphicx} 
\usepackage{natbib}  
\usepackage{caption} 
\usepackage{algorithm}
\usepackage{algpseudocode}
\usepackage{amsmath,amssymb}
\usepackage{subcaption}
\usepackage{booktabs}
\makeatletter
\newenvironment{breakablealgorithm}
  {
   \begin{center}
     \refstepcounter{algorithm}
     \hrule height.8pt depth0pt \kern2pt
     \renewcommand{\caption}[2][\relax]{
       {\raggedright\textbf{\ALG@name~\thealgorithm} ##2\par}%
       \ifx\relax##1\relax 
         \addcontentsline{loa}{algorithm}{\protect\numberline{\thealgorithm}##2}%
       \else 
         \addcontentsline{loa}{algorithm}{\protect\numberline{\thealgorithm}##1}%
       \fi
       \kern2pt\hrule\kern2pt
     }
  }{
     \kern2pt\hrule\relax
   \end{center}
  }
\makeatother

\usepackage{newfloat}
\usepackage{listings}
\DeclareCaptionStyle{ruled}{labelfont=normalfont,labelsep=colon,strut=off} 
\floatstyle{ruled}
\newfloat{listing}{tb}{lst}{}
\floatname{listing}{Listing}
\title{Modular-Cam: Modular Dynamic Camera-view Video Generation with LLM}
\author{
    Zirui Pan\textsuperscript{\rm 1},
    Xin Wang\textsuperscript{\rm 1,\rm 2}\thanks{Corresponding Author.}, Yipeng Zhang\textsuperscript{\rm 1},\\ Hong Chen\textsuperscript{\rm 1}, Kwan Man Cheng\textsuperscript{\rm 1}, Yaofei Wu\textsuperscript{\rm 3}, Wenwu Zhu$\textsuperscript{\rm 1,\rm 2}^{*}$ \\
}
\affiliations{
    \textsuperscript{\rm 1}Department of Computer Science and Technology, Tsinghua University\\
    \textsuperscript{\rm 2}Beijing National Research Center for Information Science and Technology, Tsinghua University\\
    \textsuperscript{\rm 3} Beijing University of Technology\\

    \{pzr24,zhang-yp22,h-chen20\}@mails.tsinghua.edu.cn, kcheng54@wisc.edu, 23027313@emails.bjut.edu.cn,\{xin\_wang,wwzhu\}@tsinghua.edu.cn
}

\usepackage{bibentry}

\begin{document}

\maketitle

\begin{abstract}
Text-to-Video generation, which utilizes the provided text prompt to generate high-quality videos, has drawn increasing attention and achieved great success due to the development of diffusion models recently. Existing methods mainly rely on a pre-trained text encoder to capture the semantic information and perform cross attention with the encoded text prompt to guide the generation of video. However, when it comes to complex prompts that contain dynamic scenes and multiple camera-view transformations, these methods can not decompose the overall information into separate scenes, as well as fail to smoothly change scenes based on the corresponding camera-views. To solve these problems, we propose a novel method, i.e., Modular-Cam. Specifically, to better understand a given complex prompt, we utilize a large language model to analyze user instructions and decouple them into multiple scenes together with transition actions. To generate a video containing dynamic scenes that match the given camera-views, we incorporate the widely-used temporal transformer into the diffusion model to ensure continuity within a single scene and propose CamOperator, a modular network based module that well controls the camera movements. Moreover, we propose AdaControlNet, which utilizes ControlNet to ensure consistency across scenes and adaptively adjusts the color tone of the generated video. Extensive qualitative and quantitative experiments prove our proposed Modular-Cam's strong capability of generating multi-scene videos together with its ability to achieve fine-grained control of camera movements. Generated results are available at https://modular-cam.github.io.
\end{abstract}

%

\section{Introduction}
Via training on large-scale text-image datasets, Text-to-Image (T2I) generation~\cite{rombach2022high,ramesh2022hierarchical,saharia2022photorealistic, ruiz2023dreambooth,avrahami2022blended} based on diffusion process has achieved great attention in generating high-quality images with increasing controllability. Due to the significant success of T2I models, many researchers~\cite{ho2022video,ho2022imagen,blattmann2023stable,lu2023flowzero,chen2023disenbooth,chen2024disendreamer} have made efforts to take temporal information into considerations for Text-to-Video (T2V) generation. Based on the specific text prompt, T2V models have demonstrated remarkable capability of generating videos that are smooth, photo-realistic, and semantically coherent.
\begin{figure}[t]
    \centering
    \subcaptionbox{Static camera-view. Generated by AnimateDiff.\label{fig:camera-static}}
    {
        \includegraphics[width=0.155\columnwidth]{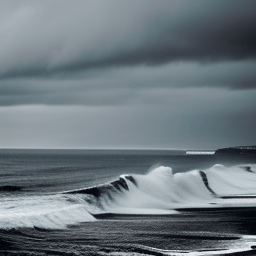}
        \includegraphics[width=0.155\columnwidth]{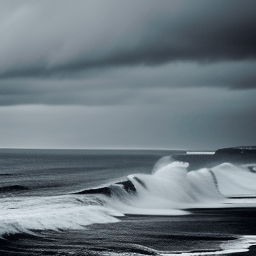}
        \includegraphics[width=0.155\columnwidth]{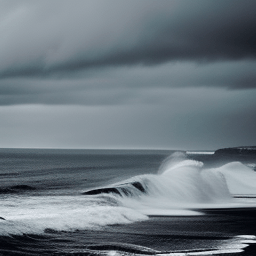}
        \includegraphics[width=0.155\columnwidth]{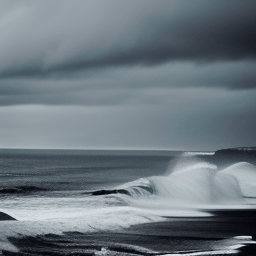}
        \includegraphics[width=0.155\columnwidth]{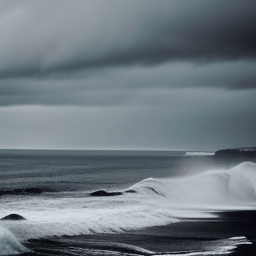}
        \includegraphics[width=0.155\columnwidth]{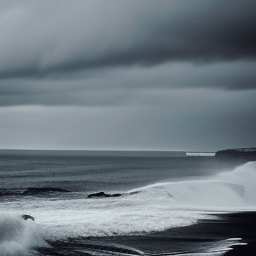}
    }
    \subcaptionbox{Inconsistentcy, mixed scenes. Generated by StreamingT2V.\label{fig:inconsistency}}
    {
        \includegraphics[width=0.155\columnwidth]{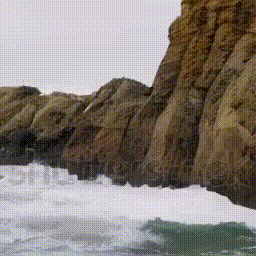}
        \includegraphics[width=0.155\columnwidth]{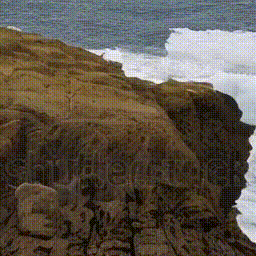}
        \includegraphics[width=0.155\columnwidth]{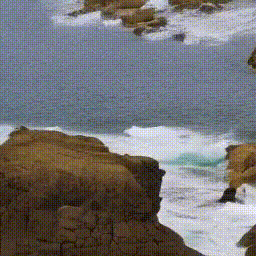}
        \includegraphics[width=0.155\columnwidth]{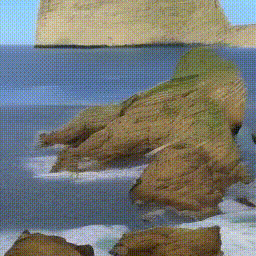}
        \includegraphics[width=0.155\columnwidth]{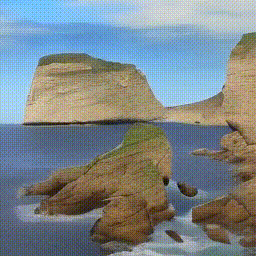}
        \includegraphics[width=0.155\columnwidth]{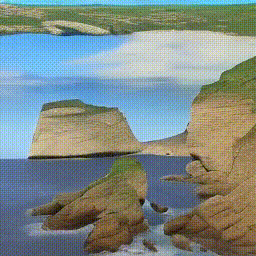}
    }
    \subcaptionbox{Generated by this work, which best follows the user instruction.\label{fig:example}}
    {
        \includegraphics[width=0.155\columnwidth]{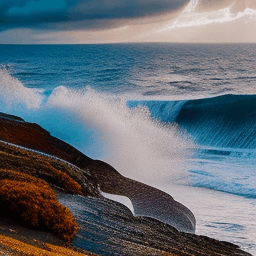}
        \includegraphics[width=0.155\columnwidth]{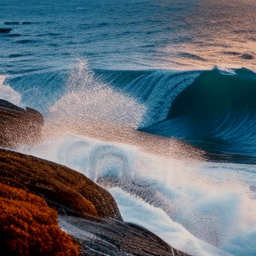}
        \includegraphics[width=0.155\columnwidth]{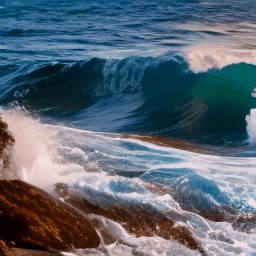}
        \includegraphics[width=0.155\columnwidth]{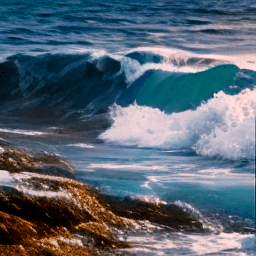}
        \includegraphics[width=0.155\columnwidth]{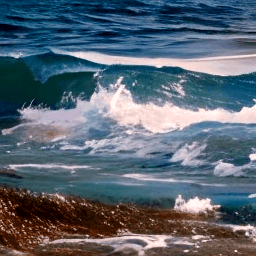}
        \includegraphics[width=0.155\columnwidth]{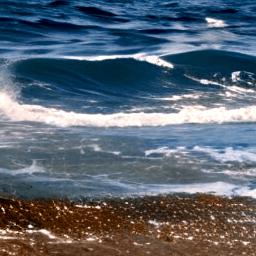}
    }
    \caption{Generated results based on instruction ``\textit{Beginning with a beach scene, the camera gradually draws in closer as waves lap against the reef. Then the camera slowly pans right and a large area of sea is revealed}". In Figure~\ref{fig:camera-static}, the video footage is almost static, while in Figure~\ref{fig:inconsistency}, the scene transitions show abrupt changes, and the scenes are mixed. Figure~\ref{fig:example} shows the results of our proposed model. 
    }
    \label{fig:current-problem}
\end{figure}

However, existing works heavily rely on the text encoder to guide the process of generation through capturing semantic information. Due to the limitation of the pre-trained encoder, it is difficult for them to understand the temporal information hidden in the complex text prompts which contain dynamic scenes changes and multiple camera-view transformations. Therefore, these models are not able to disentangle the information of different scenes, failing to sequentially generate these distinct scenes in the prescribed order of camera transformations. Consequently, the generated videos tend to only have a limited amount of motions as well as entangled scenes. Besides, current works mainly focus on generating short videos with merely 16 to 24 frames. Thus, these videos can hardly incorporate all the scenes or characterize the process of the camera transformations between adjacent ones. Although some works~\cite{henschel2024streamingt2v} adopt an autoregressive method, they fail to achieve fined control of the camera movements, suffering from color distortion and abrupt transitioning, which is destructive to the realism of the videos. 

For instance, consider an instruction shown in Figure~\ref{fig:current-problem}, which consists of three generated videos. The instruction can be decomposed into two scenes, i.e., i) \textit{Beach, waves lap against the reef} and ii) \textit{large area of sea}, with corresponding transition actions \textit{ZoomIn} and \textit{PanRight}. Figure~\ref{fig:camera-static} and Figure~\ref{fig:inconsistency} demonstrate the lack of fined control of camera movements, as well as inconsistency across multiple scenes with mixed environments and objects, respectively. Figure~\ref{fig:example} shows the results of our proposed method, which best follows the instructions. Solving the above problems is challenging since it requires a deep understanding of the instructions and thorough control of the generated content. 

To tackle the challenge, in this work we present a novel Modular-Cam framework to address the aforementioned problems. Specifically, to provide a deep understanding of the user instructions, we propose an LLM-Director which utilizes an LLM to analyze the instructions and decompose them into multiple scenes and transition actions. The obtained disentangled information is crucial for the generation of individual scenes and entire videos. Based on T2I diffusion models, we conduct a base video generator by inserting temporal transformer layers, transferring information across frames, and maintaining the continuity of the generated video within a single scene. We further propose a CamOperator module, which is a series of LoRA layers added on the base generator to ensure fine-grained control of the camera movements. For each motion pattern, i.e., \textit{ZoomIn}, \textit{PanLeft}, etc., a corresponding CamOperator module is trained. LLM will select the particular CamOperator module from the operation pool based on the transition action it acquires. 
Besides, these CamOperators can function as modular components. For complex camera-view transformations, it is not necessary to retrain each of the CamOperators but rather utilize the existing modular operators through their combinations. Benefitting from the modularity, we can easily plug them in at different situations, which greatly enhances the scalability of the model. To improve the consistency across multiple scenes, we adopt an autoregressive method and propose AdaControlNet, which introduces the ending frame of the last scene as the control information for the generation of the current scene and adaptively adjusts the color tone of the videos. Consequently, guided by the last scene, the transition between adjacent scenes will be smooth. We concatenate the video clips for each scene sequentially, deriving the final multi-scene dynamic camera-view video, which completes the end-to-end procedure.

In summary, our contributions can be listed as follows: 
\begin{itemize}
    \item We propose Modular-Cam, which is capable of generating high-quality multi-scene dynamic camera-view videos, ensuring consistency across multiple scenes, and providing a modular method to achieve fine-grained control of the contents and camera movements in the video.
    \item We propose to use LLM to parse multi-scene involved complicated user instructions, extracting scene descriptions and transition actions, and presenting an end-to-end procedure of generating multi-scene dynamic camera-view videos with modular CamOperators.
    \item We conduct extensive qualitative and quantitative experiments to verify the strong generating ability of the proposed Modular-Cam method.
\end{itemize}

\section{Related Work}

\subsection{Text-to-Video Diffusion Models}
T2V generation has become popular recently, with large-scale video datasets such as WebVid-10M \cite{bain2021frozen} that include about ten million video-text pairs collected from the Internet. Video Diffusion Model \cite{ho2022video} is one of the pioneering works in this field which extends a standard text-to-image diffusion model. However, the videos generated have poor resolution. Other works \cite{singer2022make, ho2022imagen, blattmann2023align,chen2024disenstudio,chen2023videodreamer,zhang2024scenariodiff} improve the quality through video enhancement, specifically by using spatial or temporal upsampling. On this basis, AnimateDiff \cite{guo2023animatediff} proposes using a temporal self-attention mechanism to improve frame consistency in a simple and effective way. SparseCtrl \cite{guo2023sparsectrl} further introduces a Control Encoder, adding condition images to the control information. However, many issues remain, such as style-shifting. ModelScopeT2V \cite{wang2023modelscope} ensures the consistency of generated videos and the smoothness of object motion within them by incorporating spatial-temporal awareness blocks. Nonetheless, the videos generated by the aforementioned works are still limited in length (mostly about 16 frames), making them more like animated images rather than full-fledged videos.

To generate longer videos, Text2Video-Zero \cite{khachatryan2023text2video} still relies on a text-to-image diffusion model and incorporates cross-attention from each frame to the first frame. However, as the number of frames increases, the quality of the generated videos deteriorates, and the motion in the videos remains elementary even static. 
Gen-L \cite{wang2023gen} introduces the concept of multi-text, suggesting that a long video may require multiple textual descriptions. 
FreeNoise \cite{qiu2023freenoise} adopts a method that requires no additional training, in which it manipulates the initial noise of the diffusion model so that each frame shares a small portion of it and introduces an inter-frame cross-attention mechanism. StreamingT2V \cite{henschel2024streamingt2v} employs an autoregressive approach, decomposing long video generation into the generation and stitching of multiple short videos. However, this often results in severe jitter in the visuals, and abrupt transitions may occur between adjacent short videos. Practically, when it comes to multi-scene long video generation, existing works still perform poorly in terms of scene consistency, and fail to achieve fine-grained control of camera-view transformations.

\subsection{Text-to-Video Generation Guided by LLMs}
Given the randomness of the content generated by diffusion models, it is natural to provide some control information to guide the generation process, which is already common in text-to-image diffusion models \cite{zhang2023adding}. However, for video generation, the control information becomes very complex, potentially requiring descriptions for every scene and even every frame in the video. Recently, with the continuous development of LLMs, several works \cite{lu2023flowzero, lian2023llm, long2024videodrafter, lin2023videodirectorgpt} have started to explore video generation in complex scenarios using LLMs as a breakthrough point.

LVD \cite{lian2023llm} uses LLMs to generate dynamic scene layouts to assist diffusion models in video generation. 
This idea actually comes from LayoutGPT \cite{feng2024layoutgpt}, which uses GPT \cite{achiam2023gpt} to generate a series of scene descriptions with multiple bounding boxes based on user instructions.
Similar works include FlowZero \cite{lu2023flowzero}, which also utilizes LLMs to parse instructions and generate dynamic scene layouts, introducing a self-refinement process. VideoDirectorGPT \cite{lin2023videodirectorgpt}, based on ModelScopeT2V \cite{wang2023modelscope}, incorporates scene descriptions and object layout information generated by LLMs to improve the controllability of video generation. However, the output multi-scene videos lack smooth transitions, and the generated objects cannot be accurately confined within the bounding boxes.

Other works take a different approach by using LLMs to describe scenes rather than providing layouts. This is for the reason that, in multi-scene video generation, merely providing layouts can become very complex, even for LLMs. VideoDrafter \cite{long2024videodrafter} uses LLMs to parse user instructions containing multiple scenes, generating a text description for each scene and a reference image for each entity. 
However, the multi-scene videos generated by VideoDrafter are disjointed, with abrupt transitions between adjacent scenes. Free-Bloom \cite{huang2024free} uses LLMs to generate descriptions for each keyframe in the video, employing joint denoising.
Nevertheless, this approach is also challenging in producing long multi-scene videos, as the generation quality tends to degrade when the number of frames increases.

\section{Method}
\begin{figure*}[t]
    \centering
    \includegraphics[width=\textwidth]{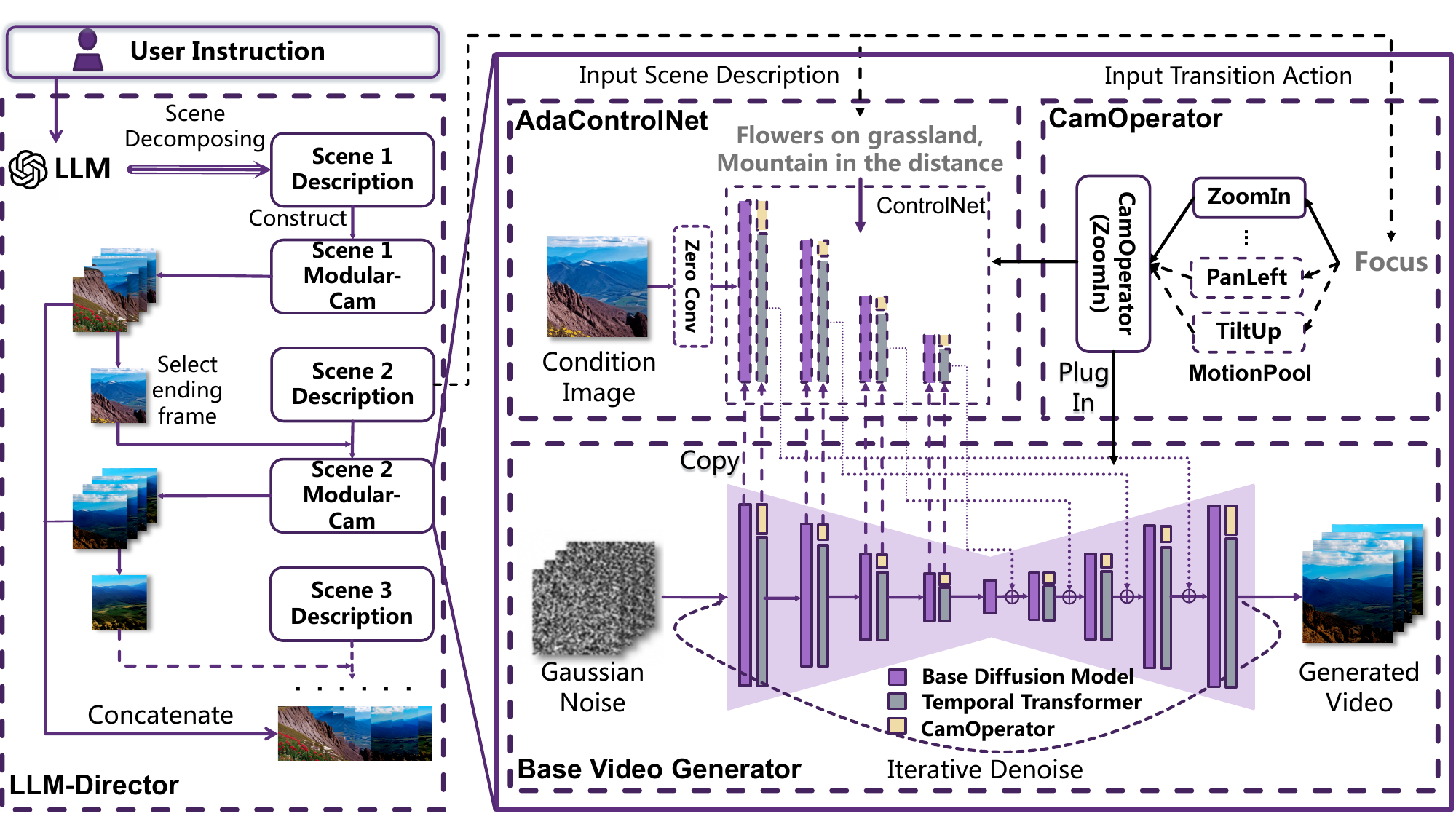}
    \caption{Framework for our proposed Modular-Cam, which contains four modules, i.e., Base Video Generator, CamOperator, AdaControlNet, and LLM-Director. First, the LLM is utilized to parse the user instruction, decomposing it into multiple scenes with descriptions. Then for each scene, a video generator is built, which has been integrated with CamOperator and AdaControlNet. LLM will identify the camera-view transformation in each scene and select from the MotionPool to plug in the appropriate CamOperator Module, which will enable the output video to follow the specific motion pattern, i.e., \textit{ZoomIn}. A condition image, that is the ending frame of the last scene, is inputted into the AdaControlNet, which will guide the generation of the current scene. Finally, the video clips for each scene are concatenated orderly to form the final multi-scene dynamic camera-view video.}
    \label{fig:framework}
\end{figure*}
In this section, we will describe our proposed Modular-Cam method. The overall framework is shown in Figure~\ref{fig:framework}. It contains a base video generator which is built upon AnimateDiff~\cite{guo2023animatediff}, a CamOperator, an AdaControlNet, and an LLM-Director. We will introduce a preliminary and give some notations we will use in this paper first and then detail each of the components in the following subsections.

\subsection{Preliminary}

\paragraph{Stable Diffusion} Stable Diffusion~\cite{rombach2022high} is a widely adopted model in T2I generation, which is open-sourced and behaves very well, thus we choose it as the base model in Modular-Cam. Stable Diffusion first utilizes a pre-trained encoder $\mathcal{E}(\cdot)$ and a pre-trained decoder $\mathcal{D}(\cdot)$ to encode and decode the image $x_0$ to and from the latent space, i.e., $z_0=\mathcal{E}(x_0)$ and $x_0^{'}=\mathcal{D}(z_0)$, respectively, performing the diffusion process in the latent space. In the forward process, the model will gradually add noise to $z_0$, until we get an approximate Gaussian noise $z_T$:
\begin{equation}
    z_t=\sqrt{\bar{\alpha}_t} z_0+\sqrt{1-\bar{\alpha}_t}\epsilon,\quad \epsilon\sim\mathcal{N}(0,1),
\end{equation}
where $t=1,2,\cdots,T$ represents the steps, and $\bar{\alpha}_t$ stands for the noise strength, $\epsilon$ is the added gaussian noise. The process of gradually adding noise is actually a Markov chain process, where we can learn to reverse it by predicting the added noise using a denoising network $\epsilon_{\theta}(\cdot)$:
\begin{equation}
    \mathcal{L}=\mathbb{E}_{\mathcal{E}(x_0),y,\epsilon\sim\mathcal{N}(0,1)}\left[\left|\left|\epsilon-\epsilon_{\theta}(z_t,t,\tau_{\theta}(y)\right|\right|_2^2\right],
\end{equation}
where $y$ represents the text description corresponding to image $x_0$, which will be encoded by a CLIP text encoder $\tau_{\theta}(\cdot)$~\cite{radford2021learning}. The text information serves as an input to guide the denoising process. To predict the specific noise, Stable Diffusion utilizes the U-Net~\cite{ronneberger2015u} which consists of symmetrical encoder and decoder. The encoder is responsible for capturing image information, while the decoder is for merging the control information with the encoded information. Each network block includes stacks of attention layers~\cite{vaswani2017attention} and residual mechanism~\cite{he2016deep}. The Base Stable Diffusion model has a large number of parameters, thus we often add LoRA~\cite{DBLP:journals/corr/abs-2106-09685} layers to finetune it, instead of tuning all the parameters. 

\paragraph{Task}
The main task of this paper is text-to-video generation, i.e., given a text prompt $p$ and the desired length $f$, generating a series of video frames $x^{1:f}$ that satisfies the requirements of the prompt. Since Stable Diffusion was originally designed for generating images, directly utilizing it to generate videos will perform poorly due to the lack of temporal information. Thus we adopt the widely used approach~\cite{guo2023animatediff} which inserts temporal transformer layers into the diffusion model to serve as our base video generator.

Suppose we have a noisy latent $z_t^{1:f}\in\mathbb{R}^{b\times c\times f\times h\times w}$, where $b$, $c$, $f$, $h$, $w$ represents the batch size, channel, frame, height and width, respectively. After each pre-trained diffusion layer, we insert a temporal transformer layer to capture temporal information of the latents. Specifically, we first reshape it to $z^{(1:f)'}_{t} \in\mathbb{R}^{f\times(b\times h\times w)\times c}$, and then we perform a self-attention along the frames axis as follows:
\begin{equation}
    \begin{aligned}
        z_t^{out}&=Attention(Query,Key,Value)\\
&=Attention(W^Q{z_t^{(1:f)}}^{'},W^K{z_t^{(1:f)}}^{'},W^V{z_t^{(1:f)}}^{'}).
    \end{aligned}
\end{equation}
Then the latents are reshaped back and incorporated into the original latents through residual connection. In this way, the temporal transformer layers will adjust the frame vectors by passing information temporally. We finetune the temporal transformer layers while keeping the pre-trained Stable Diffusion network layers fixed on large amounts of video data to learn the continuity across frames.

\subsection{CamOperator with Modular Network}\label{sec:fine-control}
After conducting the base generator, we further utilize CamOperator, which is a series of tunable LoRA layers adding to the temporal transformer layers, to control the camera movements such as \textit{PanLeft} and \textit{ZoomIn}. We finetune different sets of LoRA parameters for each motion pattern, using generated simulated video data that follow the specific pattern while keeping all the other parameters fixed:
\begin{equation}
\mathcal{W}=\mathcal{W}_{TT}+\Delta\mathcal{W}_{CO}=\mathcal{W}_{TT}+\mathcal{A}_{CO}\times\mathcal{B}^T_{CO},
\end{equation}
where $\mathcal{W}$ with subscript \textit{TT} and \textit{CO} represents parameters for the temporal transformer layer and CamOperator, respectively, and $\mathcal{A}_{CO}$ and $\mathcal{B}_{CO}$ are the low-rank matrix decompositions of $\Delta\mathcal{W}_{CO}$. We simulate the training data with such patterns using data augmentations. For example, for pattern \textit{ZoomIn}, we gradually reduce the video screen size, so that the objects in the video are slowly enlarged, thus creating a zooming-in effect. In this way, we derive CamOperator Module Pool for six basic motion patterns, that is $\textit{Motion}=\{\textit{ZoomIn},\textit{ZoomOut},\textit{PanLeft},\textit{PanRight},\textit{TiltUp},\textit{TiltDown}\}$. 

As these modules are trained individually, they can be plugged into the model independently. This modular approach greatly enhances the scalability of the model, for we can train any number of CamOperators at any time. Moreover, due to the low-rank property, these basic modules can be composed to form more complicated motions, such as \textit{PanLeft and ZoomIn}. Thus by decomposing the complicated camera movements into basic ones, we can theoretically simulate any motion pattern in the generated video.

\subsection{AdaControlNet with Randomized Blending}

In the above subsections, we build a complete model for single-scene video generation while the camera movements can be finely controlled. To generate videos with multiple scenes, we propose AdaControlNet and generate videos in an autoregressive manner, which utilizes the ending frame of the last scene as the condition image for current scene generation. However, simply using the controlnet will suffer from the abrupt transitions and the color distortions between adjacent scenes in the video. To solve the problem, we perform an adaptive pixel normalization to the controlnet which adjusts the color tone between different scenes. To further enhance the consistency between the starting frame of the video clip for the current scene and the ending frame of the last scene, we utilize a randomized blending technique inspired by~\cite{avrahami2022blended}. Thus, the generated multi-scene video can satisfy the semantic requirements while having excellent consistency.

First, we duplicate the structure and parameters of the encoder in the U-Net as our AdaControlNet. Similarly, we apply the temporal attention mechanism into the AdaControlNet so that the condition image will not only affect the starting frame but influence the rest frames as well. And we replace the input of the AdaControlNet from the concatenation of the encoded condition image and noisy latent $z_t^{1:f}$ to the encoded condition image alone, to remove the harming effects of $z_t^{1:f}$ on the AdaControlNet. We derive our training data by selecting the first frame of the video data as the condition image and finetune the AdaControlNet while keeping all the other parameters fixed. 

We find that simply using the Controlnet can ensure the consistency of objects and layouts across multiple scenes, but the color tone of the generated video may drift from the condition image, i.e., become darker or lighter. Visually, such a difference is easily recognized by the naked eye, which reduces the authenticity of the generated video. Therefore, we perform an adaptive pixel normalization which adjusts the mean and variance of the three color channels (RGB) of the generated video on the pixel-level to make it consistent with the condition image:
\begin{equation}
    frame^{ch}=\frac{frame^{ch}-frame_{mean}^{ch}}{frame_{std}^{ch}}\cdot cond_{std}^{ch}+cond_{mean}^{ch},
\end{equation}
where $frame$ and $cond$ represent a frame image or a condition image. Superscript $ch$ stands for color channel $\in\{R,G,B\}$. Additionally, we propose to use randomized blending to further unify the color tone of the generated video:
\begin{equation}\label{equ:ran}
    z_t^1=
    \left\{
    \begin{matrix}
      \begin{aligned}
        &z_t^{cond}, &if\;\texttt{random}\;(0,1) < \lambda\\
        &z_t^1, &otherwise,
      \end{aligned}
    \end{matrix}
    \right.
\end{equation}
where $z_t^1$ and $z_t^{cond}$ represent the t$^{th}$ noisy latent for the first frame of the current scene and the condition image. $\texttt{random}(0,1)$ generates a random number uniformly distributed in $(0,1)$, and $\lambda$ is a hyper-parameter controlling the intensity of randomized blending. The essence is that $z_t^1$ will receive $z_t^{cond}$ with the probability $\lambda$. We can derive that larger $\lambda$ will introduce more blending, thus the first frame will be more like the condition image, improving consistency. However, frequent blending will reduce the continuity between the first frame and the rest. On the other hand, a small $\lambda$ will encourage free generation, which improves overall continuity. In practical inference, we set $\lambda$ to 0.5.

By combining the two techniques of adaptive pixel normalization and random blending during inference, the transitions between scenes become smooth, and the multi-scene video maintains the consistency of color tone and content.

\subsection{LLM Director with Modularized Motion Selection}
Integrating AdaControlNet, we can now generate a multi-scene dynamic camera-view video. However, for a complicated multi-scene involved user instruction, the scene description or transition action may not be given directly, and the video generation model cannot automatically extract all the information. Therefore, we utilize LLM to parse user instruction, decomposing it into different scenes, and extract the transition actions between adjacent ones. LLM is like a director, guiding the video generation model in producing multi-scene videos.

Take the user instruction ``Starting with a long shot of a field and blue sky, and gradually focusing on a house in the distance. Then the camera moves to the left, and large fields appear, then the house moves out of view" as an example, which has two scenes. We design the prompt:
\begin{itemize}
    \item "\textit{Extract the scenes that appear in the given text in order, and identify the transition actions between adjacent scenes. The scene description should contain rich information. You should pick the transition action from [Zoom In, Zoom Out, Pan Left, Pan Right, Tilt Up, Tilt Down]}". 
\end{itemize}

LLM will analyze the user instruction, decomposing it into the two scenes and output to the specific format for the video generation model to receive:
\begin{equation}
  \begin{matrix}
    \begin{aligned}
      [&Scene 1: \text{``field and blue sky, house in the distance"}, \\&Action: \text{ZoomIn}]\\
      [&Scene 2: \text{``large fields"}, Action: \text{PanLeft}].
    \end{aligned}
  \end{matrix}
\end{equation}

For \textit{Scene1}, our video generator will select and integrate \textit{ZoomIn} CamOperator Module to generate a single scene video. And for \textit{Scene2}, our video generator will select the last frame of \textit{Scene1} as a condition image and integrate \textit{PanLeft} CamOperator Module for the generation. The final video is derived by concatenating video clips for \textit{Scene1} and \textit{Scene2}.

Note that although LLM is not directly involved in specific video generation, it still plays an indispensable role, just like what an excellent director can bring to a film. If we input the user instruction directly to the video generation model without LLM parsing, the scenes in the generated video will be mixed. 

Thus, we get Modular-Cam, capable of generating dynamic camera-view transformations and multi-scene long videos based on complex user instructions.

\section{Experiment}
In this section, we first detail on the specific setting of the training and testing of our proposed Modular-Cam, and conduct extensive quantitative and qualitative experiments to demonstrate the strong generating ability of our model. We further conduct some ablation studies to verify the effectiveness of each module.
\subsection{Experiment Setup}
We use the large-scale public video dataset WebVid-10M~\cite{bain2021frozen} as our training set to train the newly inserted temporal transformer layers. For CamOperator, we simulate and generate about 50 videos with specific motion patterns to finetune the LoRA layers. For AdaControlNet, we select 100,000 videos from WebVid-10M as the training set, with the starting frame as the condition image. The whole training procedure can be found in the Appendix.

In the inference stage, since our work mainly focuses on multi-scene dynamic camera-view video generation, current instruction sets, which mostly contain simple single-scene instructions, cannot satisfy our requirements. Therefore, in quantitative comparison, we adopt a self-generated dataset, which contains 1000 multi-scene involved instructions. We use ChatGPT3.5-turbo~\cite{achiam2023gpt} to auto-generate the dataset, ensuring that each instruction has at least two scenes with guidance on camera-view transformations. 

We compare Modular-Cam with the baselines in terms of five metrics, i.e., \textit{Motion Smoothness}(MS), \textit{Dynamic Degree}(DD), \textit{Imaging Quality}(IQ), CLIP Metric and \textit{User Rank}(UR). The detailed information of the employed metrics can be found in the appendix.


\subsection{Main Results}
\begin{table}[t]
	\centering
 \setlength{\tabcolsep}{1mm}
    \begin{tabular}{cccccc}
    \toprule  
    
    \textbf{Model} & \textbf{MS}($\uparrow$) & \textbf{DD}($\uparrow$) & \textbf{IQ}($\uparrow$) & \textbf{CLIP}($\uparrow$) & \textbf{UR}($\downarrow$)\\
    \midrule 
    AnimateDiff & 0.983 & 0.329 & \textbf{0.622} & 0.171 &3.8\\
    FreeNoise & \underline{0.986} & 0.302 & \underline{0.618} & 0.171 &3.7\\
    SparseCtrl & 0.952 & 0.677 & 0.524 & 0.208 & 3.5\\
    StreamingT2V & 0.974 & \underline{0.907} & 0.350 & \underline{0.224} &\underline{2.3}\\
    \midrule
    Modular-Cam & \textbf{0.988} & \textbf{0.994} & 0.546 & \textbf{0.232} &\textbf{1.7}\\
    \bottomrule 
    \end{tabular}
    \caption{Quantitative comparison between Modular-Cam and other baselines, 
    where MS, DD and IQ stands for \textit{Motion Smoothness}, \textit{Dynamic Degree} and \textit{Imaging Quality}, respectively, and UR represents \textit{User Rank}, a user evaluation metric. The top and second top performances have been bolded or underlined, respectively. $\uparrow$ represents that the higher the metric, the better, while $\downarrow$ represents the opposite.}
 \label{tab:quantitative}
\end{table}

\begin{figure}[t]
    \centering
    \subcaptionbox{AnimateDiff generated results.\label{fig:qualitative-animatediff}}
    {
        \includegraphics[width=0.155\linewidth]{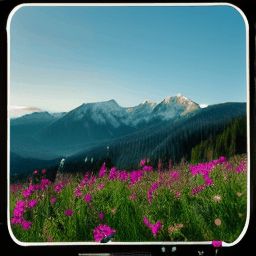}
        \includegraphics[width=0.155\linewidth]{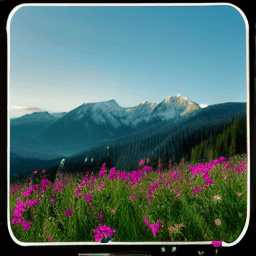}
        \includegraphics[width=0.155\linewidth]{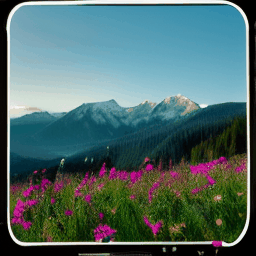}
        \includegraphics[width=0.155\linewidth]{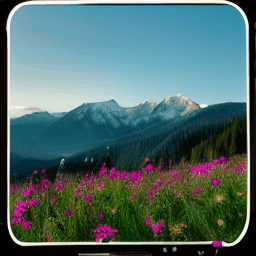}
        \includegraphics[width=0.155\linewidth]{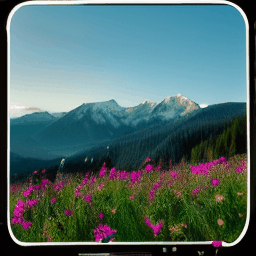}
        \includegraphics[width=0.155\linewidth]{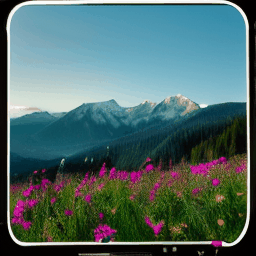}
    }
    \subcaptionbox{FreeNoise generated results.\label{fig:qualitative-freenoise}}
    {
        \includegraphics[width=0.155\linewidth]{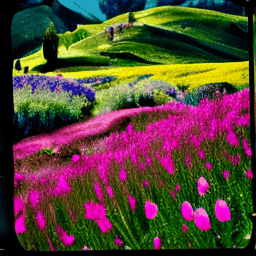}
        \includegraphics[width=0.155\linewidth]{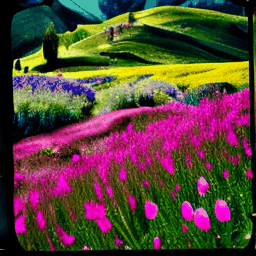}
        \includegraphics[width=0.155\linewidth]{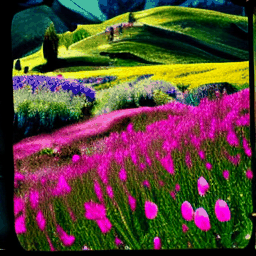}
        \includegraphics[width=0.155\linewidth]{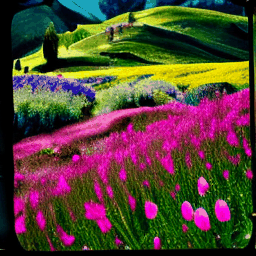}
        \includegraphics[width=0.155\linewidth]{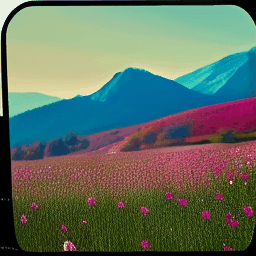}
        \includegraphics[width=0.155\linewidth]{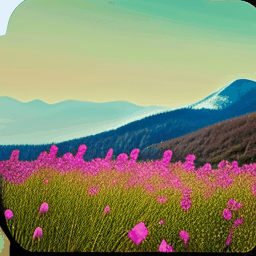}
    }
    \subcaptionbox{SparseCtrl generated results.\label{fig:qualitative-sparsectrl}}
    {
        \includegraphics[width=0.155\linewidth]{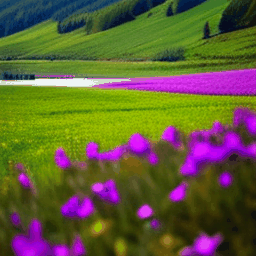}
        \includegraphics[width=0.155\linewidth]{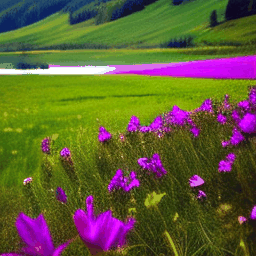}
        \includegraphics[width=0.155\linewidth]{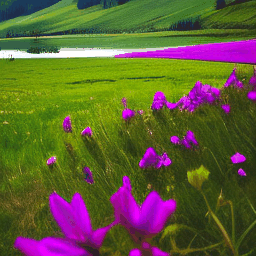}
        \includegraphics[width=0.155\linewidth]{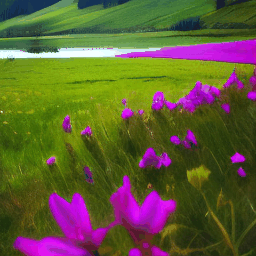}
        \includegraphics[width=0.155\linewidth]{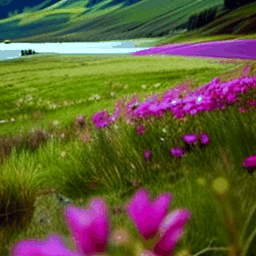}
        \includegraphics[width=0.155\linewidth]{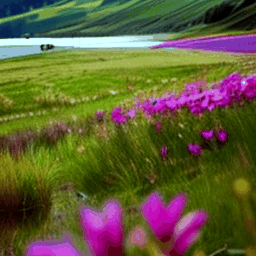}
    }
    \subcaptionbox{StreamingT2V generated results.\label{fig:qualitative-streamingt2v}}
    {
        \includegraphics[width=0.155\linewidth]{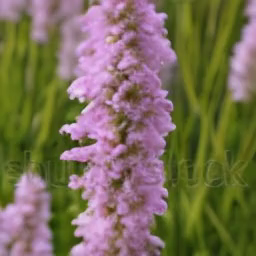}
        \includegraphics[width=0.155\linewidth]{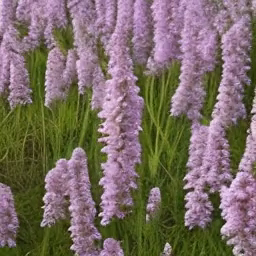}
        \includegraphics[width=0.155\linewidth]{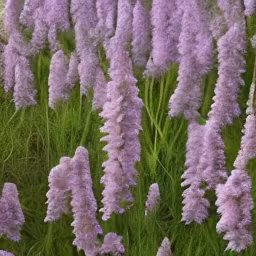}
        \includegraphics[width=0.155\linewidth]{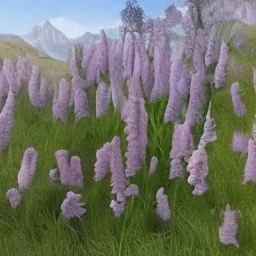}
        \includegraphics[width=0.155\linewidth]{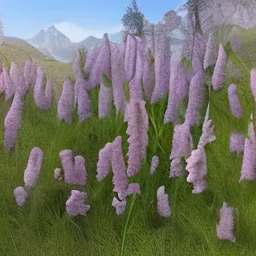}
        \includegraphics[width=0.155\linewidth]{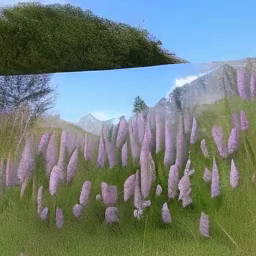}
    }
    \subcaptionbox{Modular-Cam generated results.\label{fig:qualitative-MoDPM-Director}}
    {
        \includegraphics[width=0.155\linewidth]{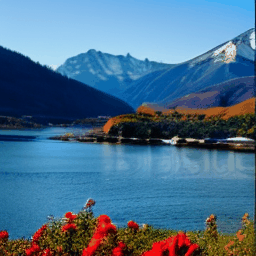}
        \includegraphics[width=0.155\linewidth]{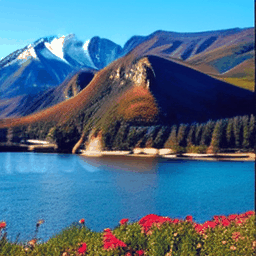}
        \includegraphics[width=0.155\linewidth]{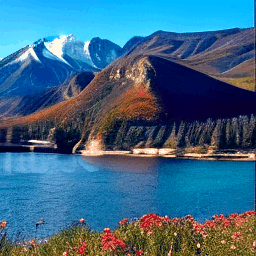}
        \includegraphics[width=0.155\linewidth]{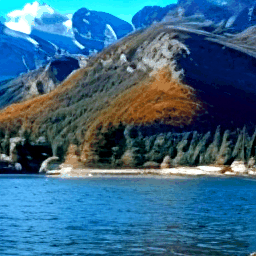}
        \includegraphics[width=0.155\linewidth]{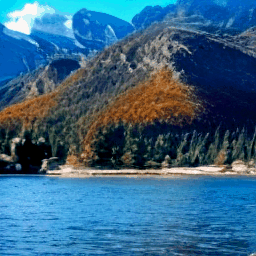}
        \includegraphics[width=0.155\linewidth]{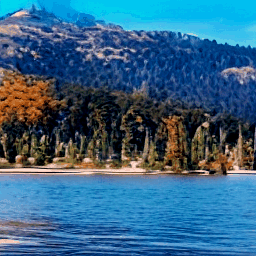}
    }
    \caption{Qualitative comparison between Modular-Cam and other baselines. We select several intermediate frames of the whole video for the convenience of presentation.}
    \label{fig:qualitative-result}
\end{figure}

We conduct quantitative and qualitative comparisons and demonstrate the results in Table~\ref{tab:quantitative} and Figure~\ref{fig:qualitative-result}. 
In quantitative comparison, we can observe that Modular-Cam outperforms the other baselines in most of the metrics, no matter computed or manually evaluated, and is only slightly lower than AnimateDiff and FreeNoise under \textit{IQ}, which may be attributed to the trade-off between content richness and quality, since Modular-Cam generates a complicated video with multi-scenes and camera-view transformations, and is relatively harder to maintain the same-level quality. In terms of \textit{Dynamic Degree}, Modular-Cam is much higher than the other baselines, validating its superior dynamic camera-view generating ability. StreamingT2V also focuses on multi-scene video generation and performs similarly to Modular-Cam in terms of \textit{CLIP Metric} and \textit{MS}. However, it falls behind in \textit{UR} and \textit{DD}, and drops drastically on \textit{IQ}.

In qualitative comparison, we evaluate Modular-Cam and other baselines based on the same user instruction, that is \textit{``Starting with a close up shot of the flowers in the meadow, the camera slowly moves to the right to focus on the mountain peaks in the distance and gradually draws in closer. The camera then continues to move to the right as the mountain and the lake mirror each other"}. The results are shown in Figure~\ref{fig:qualitative-result}. The instruction can be decomposed into three scenes, i.e., \textit{close up shot of the flowers in the meadow}, \textit{gradually focus on the mountain in the distance} and \textit{mountain and lake mirror each other}, with transition actions \textit{PanRight}, \textit{ZoomIn} and \textit{PanRight}, respectively. We can observe that the videos generated by AnimateDiff, FreeNoise and SparseCtrl are almost static in motion, especially in results generated by FreeNoise, the frames change completely in the later stage, potentially affected by the \textit{ZoomIn} instruction, while StreamingT2V displays abrupt transitions, with mixed scenes in the end, i.e., mountain and flower. On the other hand, the video outputted by Modular-Cam best follows the user instructions, generating all the objects correctly and performing the right camera-view transformations. Additional results can be found in the Appendix.

\subsection{Ablation Studies}
In this subsection, we validate the effectiveness of the proposed modules through a series of ablation studies.

\begin{figure}[t]
    \centering
    \begin{subfigure}{.24\columnwidth}
        \centering
        \includegraphics[width=\linewidth]{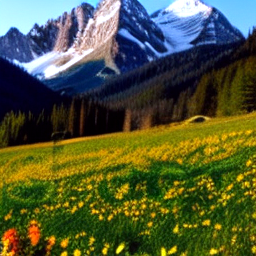}
        \caption{Condition.}\label{fig:meanstd-1}
    \end{subfigure}
    \begin{subfigure}{.24\columnwidth}
        \centering
        \includegraphics[width=\linewidth]{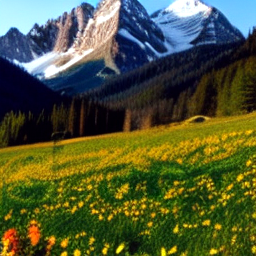}
        \caption{Original.}\label{fig:meanstd-gen-1}
    \end{subfigure}
    \begin{subfigure}{.24\columnwidth}
        \centering
        \includegraphics[width=\linewidth]{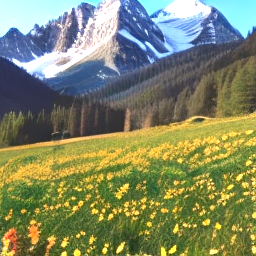}
        \caption{w/o. RB.}\label{fig:meanstd-gen-2}
    \end{subfigure}
    \begin{subfigure}{.24\columnwidth}
        \centering
        \includegraphics[width=\linewidth]{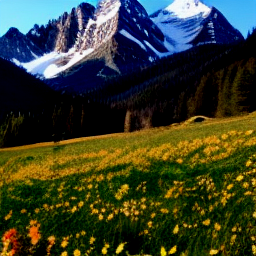}
        \caption{w/o. Ad.}\label{fig:meanstd-gen-3}
    \end{subfigure}

    \begin{subfigure}{.242\columnwidth}
        \centering
        \includegraphics[width=\linewidth]{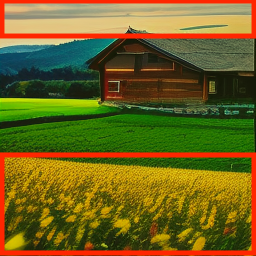}
        \caption{Condition.}\label{fig:meanstd-2}
    \end{subfigure}
    \hfill
    \begin{subfigure}{.242\columnwidth}
        \centering
        \includegraphics[width=\linewidth]{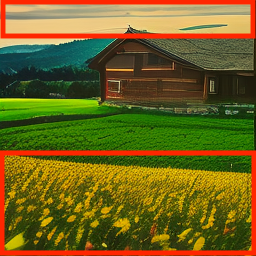}
        \caption{Original.}\label{fig:meanstd-gen-4}
    \end{subfigure}
    \hfill
    \begin{subfigure}{.242\columnwidth}
        \centering
        \includegraphics[width=\linewidth]{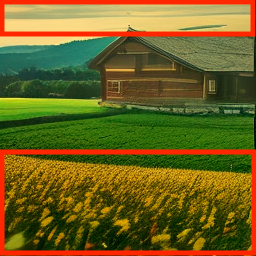}
        \caption{w/o. RB.}\label{fig:meanstd-gen-5}
    \end{subfigure}
    \hfill
    \begin{subfigure}{.242\columnwidth}
        \centering
        \includegraphics[width=\linewidth]{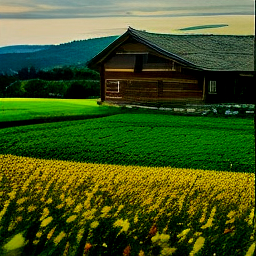}
        \caption{w/o. Ad.}\label{fig:meanstd-gen-6}
    \end{subfigure}
    \caption{Ablation study on adjusting the color tone, where \textit{Original} represents the generated results of Modular-Cam, and \textit{RB} and \textit{Ad} stands for \textit{Randomized blending} and \textit{Adaptive pixel normalization}, respectively. We remove the two techniques and display the first frame of each generated video compared with the condition image, where areas with color tone shifting are marked with red frames.}\label{fig:meanstd}
\end{figure}

\paragraph{AdaControlNet}
Due to the misalignment of data distributions in training and inference, color tone shifting usually occurs in the generated results of diffusion models~\cite{song2020improved}. Therefore, we propose to use \textit{Adaptive pixel normalization} and \textit{Randomized blending} to mitigate the shifting problem. From the results in Figure~\ref{fig:meanstd}, we can observe that without \textit{Randomized blending}, the frames become slightly whiter or darker, as has been marked with the red frame, while without \textit{Adaptive pixel normalization}, the frames overall obviously turn darker. The change of color tone, even the smallest, may possibly be recognized by the naked eye, thus destructing the general realism. Utilizing the two techniques, the generated results of Modular-Cam showcase the most aligned color tone, which enhances the authenticity of the generated video.

\paragraph{LLM-Director}
\begin{figure}[t]
    \centering
    \subcaptionbox{Video generated with LLM decomposing multi-scene.\label{fig:analysis-llm-yes}}
    {
        \includegraphics[width=0.155\columnwidth]{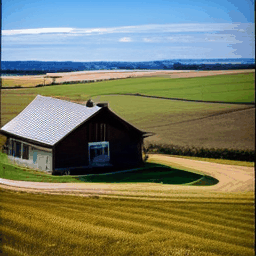}
        \includegraphics[width=0.155\columnwidth]{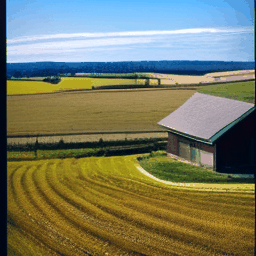}
        \includegraphics[width=0.155\columnwidth]{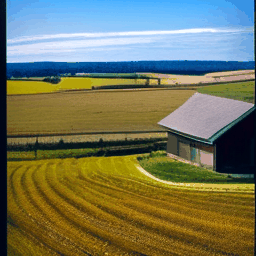}
        \includegraphics[width=0.155\columnwidth]{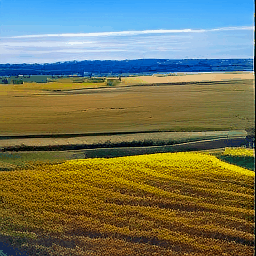}
        \includegraphics[width=0.155\columnwidth]{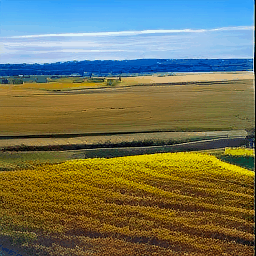}
        \includegraphics[width=0.155\columnwidth]{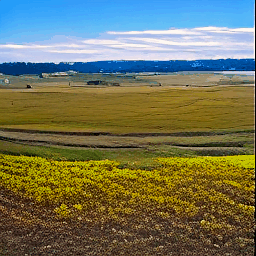}
    }
    \subcaptionbox{Video generated directly using the multi-scene user instruction.\label{fig:analysis-llm-no}}
    {
        \includegraphics[width=0.155\columnwidth]{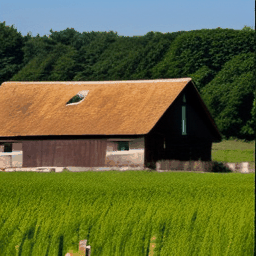}
        \includegraphics[width=0.155\columnwidth]{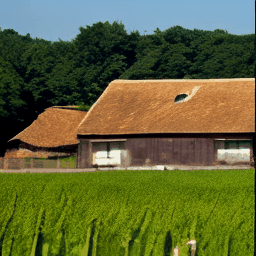}
        \includegraphics[width=0.155\columnwidth]{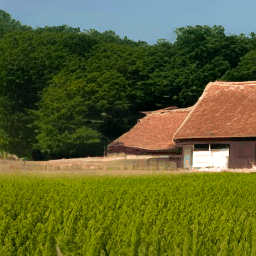}
        \includegraphics[width=0.155\columnwidth]{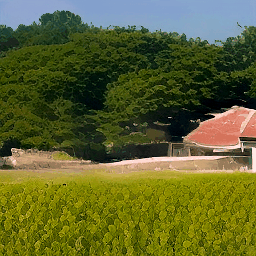}
        \includegraphics[width=0.155\columnwidth]{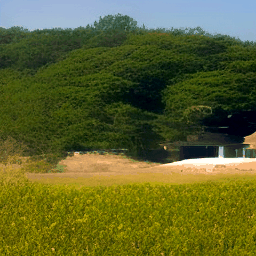}
        \includegraphics[width=0.155\columnwidth]{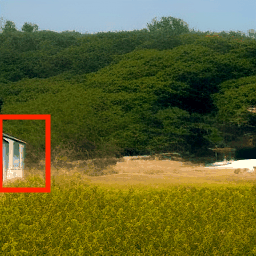}
    }
    \caption{Ablation study on LLM decomposing user instruction. In Modular-Cam, the instruction is first parsed and decomposed by LLM then be fed to the video generator, while in Figure~\ref{fig:analysis-llm-no}, we display the generated result of video generator directly utilizing the multi-scene involved instruction.}
    \label{fig:analysis-llm}
\end{figure}
To illustrate the important role of LLM decomposing multi-scene involved instructions, we design a simple prompt ``\textit{Beginning with a scene of fields and houses, the camera gradually moves to the left, the houses move out of view, and large fields appear}", which consists of only two scenes, with transition action \textit{PanLeft}. We parse it with LLM and obtain the decomposed two scenes as \textit{fields and house} and \textit{large fields}, in which we can find that the LLM has understood the instruction to remove the object \textit{house} out of the second scene, avoiding confusing the video generator. We compare it with the generated video of directly utilizing the undecomposed instruction, i.e., the description for each scene is the original multi-scene user instruction. The results are in Figure~\ref{fig:analysis-llm}. We can observe that in Figure~\ref{fig:analysis-llm-no}, mixed scenes occur in scene 2, where objects similar to \textit{house} appear again in the scene, as has been marked with the red frame, which is contrary to the \textit{move out of view} instruction, while in Figure~\ref{fig:analysis-llm-yes} only large fields remain, showcasing the strong comprehension capability brought by the LLM.

\begin{figure}[t]
    \begin{minipage}{.25\columnwidth}
        \begin{subfigure}{\textwidth}
            \centering
            \includegraphics[width=\linewidth]{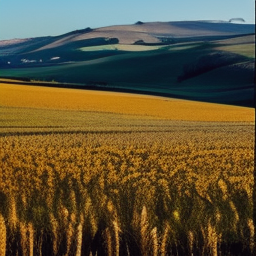}
            \caption{Condition.}\label{fig:blending-0}
        \end{subfigure}
    \end{minipage}
    \hspace{1mm}
    \begin{minipage}{.7\columnwidth}
        \begin{subfigure}{.232\textwidth}
            \centering
            \includegraphics[width=\linewidth]{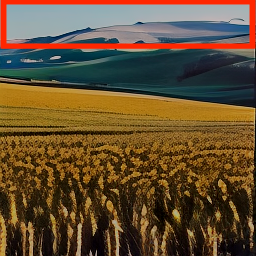}
            \captionsetup{justification=centering}
            \caption{$\lambda=0.25$\\$1^{\text{st}}$ \text{frame}.}\label{fig:blending-1-1}
        \end{subfigure}
        \hspace{0mm}
        \begin{subfigure}{.232\textwidth}
            \centering
            \includegraphics[width=\linewidth]{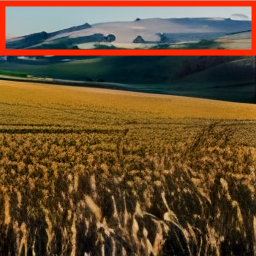}
            \captionsetup{justification=centering}
            \caption{$\lambda=0.25$\\$2^{\text{nd}}$ \text{frame}.}\label{fig:blending-1-2}
        \end{subfigure}
        \hspace{0mm}
        \begin{subfigure}{.232\textwidth}
            \centering
            \includegraphics[width=\linewidth]{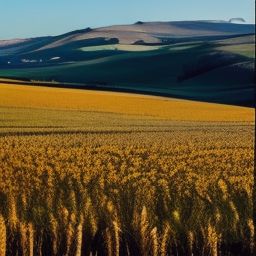}
            \captionsetup{justification=centering}
            \caption{$\lambda=0.5$\\$1^{\text{st}}$ \text{frame}.}\label{fig:blending-2-1}
        \end{subfigure}
        \hspace{0mm}
        \begin{subfigure}{.232\textwidth}
            \centering
            \includegraphics[width=\linewidth]{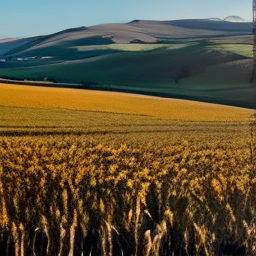}
            \captionsetup{justification=centering}
            \caption{$\lambda=0.5$\\$2^{\text{nd}}$ \text{frame}.}\label{fig:blending-2-2}
        \end{subfigure}
        
        \begin{subfigure}{.232\textwidth}
            \centering
            \includegraphics[width=\linewidth]{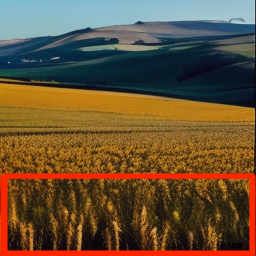}
            \captionsetup{justification=centering}
            \caption{$\lambda=0.75$\\$1^{\text{st}}$ \text{frame}.}\label{fig:blending-3-1}
        \end{subfigure}
        \hspace{0mm}
        \begin{subfigure}{.232\textwidth}
            \centering
            \includegraphics[width=\linewidth]{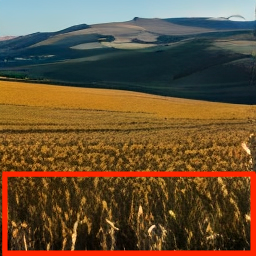}
            \captionsetup{justification=centering}
            \caption{$\lambda=0.75$\\$2^{\text{nd}}$ \text{frame}.}\label{fig:blending-3-2}
        \end{subfigure}
        \hspace{0mm}
        \begin{subfigure}{.232\textwidth}
            \centering
            \includegraphics[width=\linewidth]{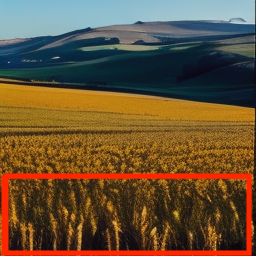}
            \captionsetup{justification=centering}
            \caption{$\lambda=1.0$\\$1^{\text{st}}$ \text{frame}.}\label{fig:blending-4-1}
        \end{subfigure}
        \hspace{0mm}
        \begin{subfigure}{.232\textwidth}
            \centering
            \includegraphics[width=\linewidth]{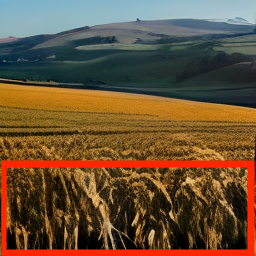}
            \captionsetup{justification=centering}
            \caption{$\lambda=1.0$\\$2^{\text{nd}}$ \text{frame}.}\label{fig:blending-4-2}
        \end{subfigure}
    \end{minipage}
    \caption{Sensitivity analysis on the $\lambda$ introduced in randomized blending. Here we only display the $1^{\text{st}}$ and $2^{\text{nd}}$ frame of the generated results for specific value of $\lambda$. We mark the areas where inconsistent color tone or transition gap occurs with red frame.}\label{fig:blending}
    \vspace{-3mm}
\end{figure}
\subsection{Parameter Sensitivity}
In Equation~\ref{equ:ran}, we introduce a hyper-parameter $\lambda$ to control the intensity of randomized blending. We determine the optimal value of $\lambda$ through sensitivity analysis. Since randomized blending only directly impact on the $1^{\text{st}}$ frame, we can judge the continuity and consistency of the video by comparing the $1^{\text{st}}$ frame with the condition image and the $2^{\text{nd}}$ frame, which represents the rest of the video, respectively. From Figure~\ref{fig:blending}, we can observe that small $\lambda$ reduces consistency with the condition image in terms of color tone, while large $\lambda$ results in transition gap between the $1^{st}$ frame and the rest of the video, where misaligned shapes and positions of objects occur, which confirms our presumption. Experimentally, we find that $\lambda=0.5$ achieves the best balance between the continuity within a single scene and the consistency across multiple scenes.

\section{Conclusion}
In this work, we present a novel method called Modular-Cam, which is able to generate multi-scene dynamic camera-view video, overcoming the limitations of existing works, which either output videos that are almost static, without much motion dynamics, or produce severe gaps between adjacent scenes. We propose three modules to address these problems, namely CamOperator, AdaControlNet, and LLM-Director, to enhance the consistency across multiple scenes and provide fine-grained control of camera movements, where we utilize modular network to learn each motion pattern and take advantage of LLM's understanding capacity to guide the video generation. Extensive experiments verify the strong generating ability of Modular-Cam and the effectiveness of each proposed module.

\section{Acknowledgements}
This work was supported by the National Key Research and Development Program of China No.2023YFF1205001, National Natural Science Foundation of China No. 62222209, Beijing National Research Center for Information Science and Technology under Grant No. BNR2023TD03006, BNR2023RC01003, and Beijing Key Lab of Networked Multimedia.
\bibliography{aaai25}

\section{Technical Appendix for\\Modular-Cam: Modular Dynamic Camera-view Video Generation with LLM}

In this technical appendix, we first list the details of our implementation as well as the employed evaluation metrics and training procedure of Modular-Cam for the convenience of reproduction. Then we give additional experimental results, including additional ablation studies on LLM and more qualitative experimental results, and then discuss the limitation for further investigation. Generated results in terms of video can be found in https://modular-cam.github.io.

\section{Implementation Details}
In this section, we give the implementation details for training. We set the resolution for each frame in the video to 256$\times$256, and fix the length of a single video clip to 16 frames, while the number of clips in a single scene or in the whole generated video are not limited. The base diffusion model we used is Stable Diffusion v1-5\footnote{https://huggingface.co/runwayml/stable-diffusion-v1-5}, which has loaded the DreamBooth~\cite{ruiz2023dreambooth} LoRA\footnote{https://civitai.com/models/4201?modelVersionId=130072}~\cite{DBLP:journals/corr/abs-2106-09685}. The LLM used in LLM-Director is chatgpt3.5-turbo-1106. The rank of the LoRA layers in CamOperator is set to 4. We choose the Adam~\cite{kingma2014adam} optimizer and set the learning rate to 1e-4. The training is conducted on Nvidia A100 40G GPU under linux environment, using the deep learning framework PyTorch.

We have employed \textit{Motion Smoothness}, \textit{Dynamic Degree}, \textit{Imaging Quality}, \textit{CLIP score} and \textit{User Rank} as metrics. The first three metrics are supported by VBench~\cite{huang2024vbench}, an open-source video evaluation benchmark. For\textit{ CLIP score}, we calculate the score by averaging the cosine similarity between each video frame to the CLIP encoded user instruction, which simulates the compliance of the generated video with the given instruction. To better evaluate the performance of each method, we conduct a user study and calculate the \textit{User Rank} metric. Specifically, we ask 15 participants to rank 20 sets of videos which are generated by Modular-Cam and baselines using the same user instruction, based on intuitive feeling, video quality, etc. We then average the scores for each model as the final \textit{User Rank}.

\section{Additional Ablation Study on LLM}
In this section, we conduct additional ablation studies to verify the effectiveness of LLM. We quantitatively compare the performance of Modular-Cam using different types of prompts, i.e., the original prompts and the LLM decomposed prompts. We follow the same experimental settings as in the main paper. The results can be seen in Table~\ref{tab:quantitative-llm}.

\begin{table}[h]
	\centering
 \setlength{\tabcolsep}{1mm}
    \begin{tabular}{ccccc}
    \toprule  
    
    \textbf{Model} & \textbf{MS}($\uparrow$) & \textbf{DD}($\uparrow$) & \textbf{IQ}($\uparrow$) & \textbf{CLIP}($\uparrow$) \\
    \midrule 
    Modular-Cam(w/o. LLM) & 0.950 & 0.771 & 0.511 & 0.221 \\
    Modular-Cam & \textbf{0.988} & \textbf{0.994} & \textbf{0.546} & \textbf{0.232} \\
    \bottomrule 
    \end{tabular}
    \caption{Quantitative comparison between using the original prompts and the LLM decomposed prompts. $\uparrow$ represents that the higher the metric, the better, while $\downarrow$ represents the opposite.}
 \label{tab:quantitative-llm}
\end{table}

From the results we can observe that after LLM director decomposing the multiple scenes, the qualities of the generated videos are indeed improved, especially on the Dynamic Degree metric, which is lifted from 0.771 to 0.994, since the decomposed scene will assist the video generator to handle the complex prompt.

\begin{figure*}[h]
    \centering
    \subcaptionbox{\textit{Beginning with a beach scene, the camera gradually \textbf{draws in} closer as waves lap against the reef. Then the camera slowly \textbf{pans right} and a large area of sea is revealed}.}
    {
        \includegraphics[width=0.115\linewidth]{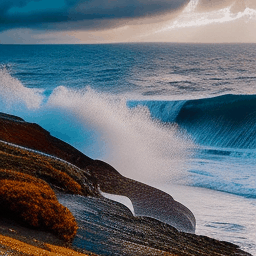}
        \includegraphics[width=0.115\linewidth]{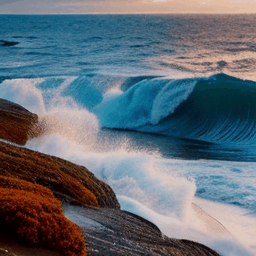}
        \includegraphics[width=0.115\linewidth]{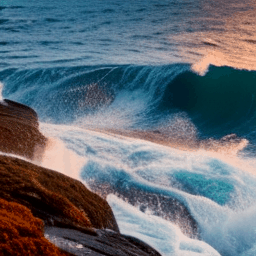}
        \includegraphics[width=0.115\linewidth]{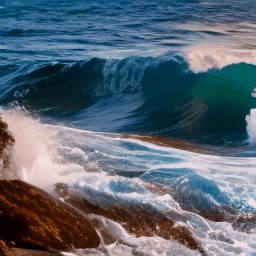}
        \includegraphics[width=0.115\linewidth]{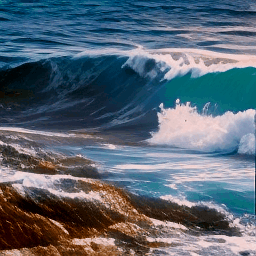}
        \includegraphics[width=0.115\linewidth]{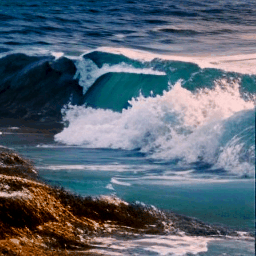}
        \includegraphics[width=0.115\linewidth]{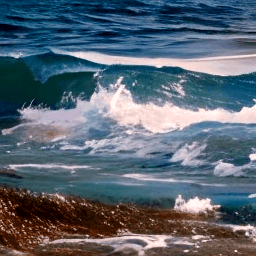}
        \includegraphics[width=0.115\linewidth]{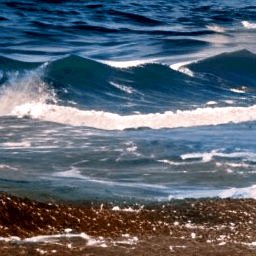}
    }
    \subcaptionbox{\textit{Starting with a close up shot of the flowers in the meadow, the camera slowly \textbf{moves to the right} to focus on the mountain peaks in the distance and gradually \textbf{draws in closer}. The camera then continues to \textbf{move to the right} as the mountain and the lake mirror each other}.}
    {
        \includegraphics[width=0.115\linewidth]{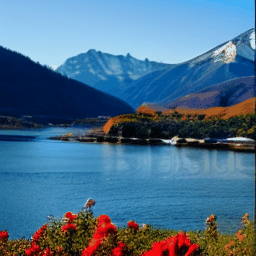}
        \includegraphics[width=0.115\linewidth]{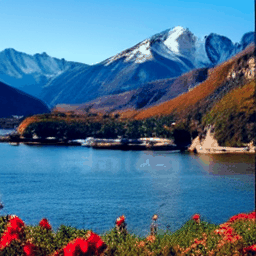}
        \includegraphics[width=0.115\linewidth]{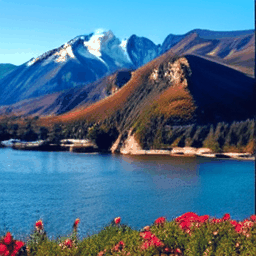}
        \includegraphics[width=0.115\linewidth]{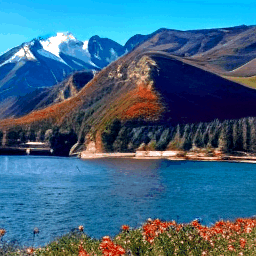}
        \includegraphics[width=0.115\linewidth]{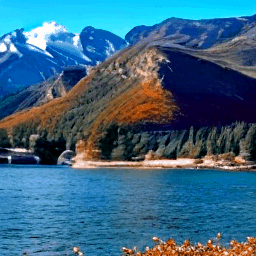}
        \includegraphics[width=0.115\linewidth]{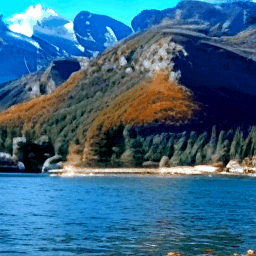}
        \includegraphics[width=0.115\linewidth]{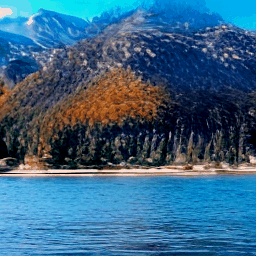}
        \includegraphics[width=0.115\linewidth]{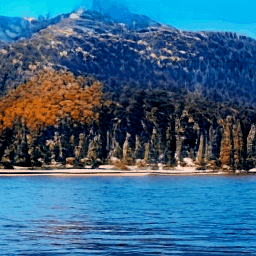}
    }
    \subcaptionbox{\textit{The camera \textbf{pans left} to reveal a large area of desert, then the camera \textbf{advances} to give a better view of the dunes in the distance. Finally, the camera \textbf{tilts down} to focus on the sand at the feet}.}
    {
        \includegraphics[width=0.115\linewidth]{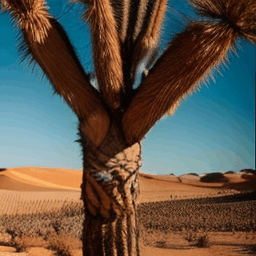}
        \includegraphics[width=0.115\linewidth]{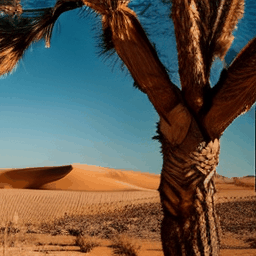}
        \includegraphics[width=0.115\linewidth]{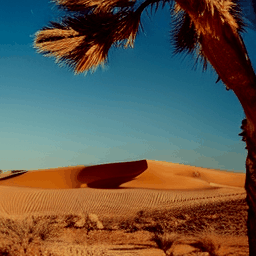}
        \includegraphics[width=0.115\linewidth]{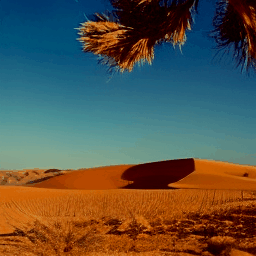}
        \includegraphics[width=0.115\linewidth]{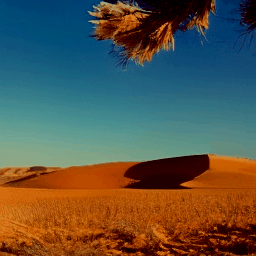}
        \includegraphics[width=0.115\linewidth]{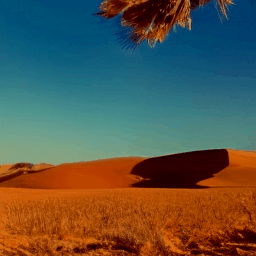}
        \includegraphics[width=0.115\linewidth]{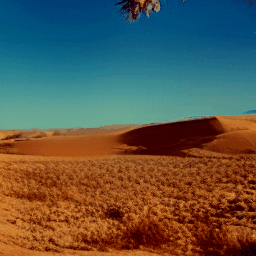}
        \includegraphics[width=0.115\linewidth]{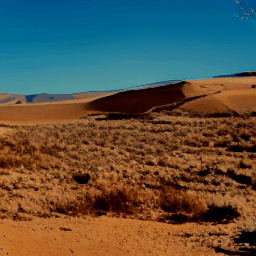}
    }

    \subcaptionbox{\textit{Starting with a distant shot of a field with blue sky, the camera gradually \textbf{focuses on} a house in the distance. The camera then gradually \textbf{pans right} to reveal a large field and the house moves out of view}.}
    {
        \includegraphics[width=0.115\linewidth]{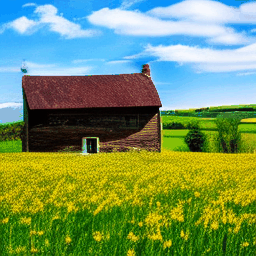}
        \includegraphics[width=0.115\linewidth]{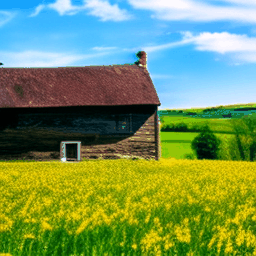}
        \includegraphics[width=0.115\linewidth]{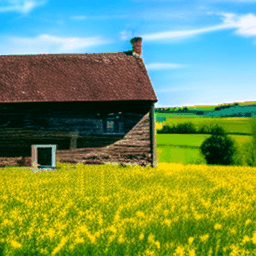}
        \includegraphics[width=0.115\linewidth]{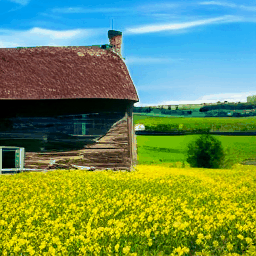}
        \includegraphics[width=0.115\linewidth]{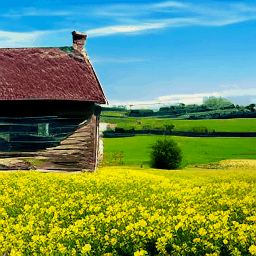}
        \includegraphics[width=0.115\linewidth]{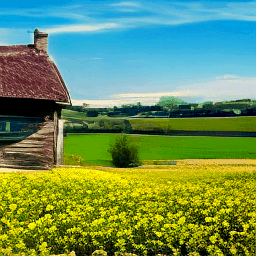}
        \includegraphics[width=0.115\linewidth]{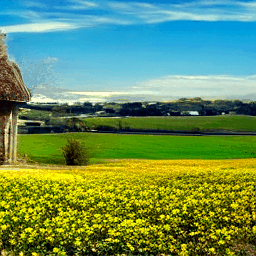}
        \includegraphics[width=0.115\linewidth]{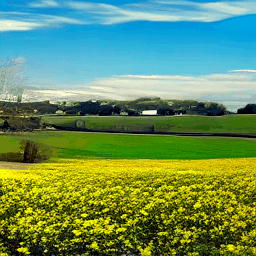}
    }
    \subcaptionbox{\textit{Starting in a submarine world, the camera slowly \textbf{moves left} to reveal a variety of marine life. Then the camera \textbf{moves forward}, aiming at a large area of water plants in the distance}.}
    {
        \includegraphics[width=0.115\linewidth]{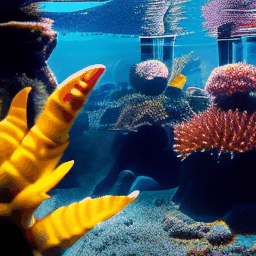}
        \includegraphics[width=0.115\linewidth]{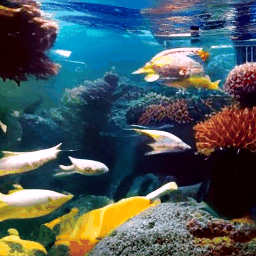}
        \includegraphics[width=0.115\linewidth]{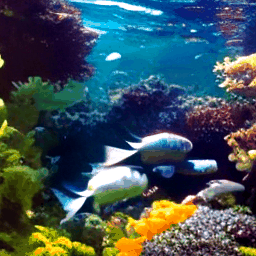}
        \includegraphics[width=0.115\linewidth]{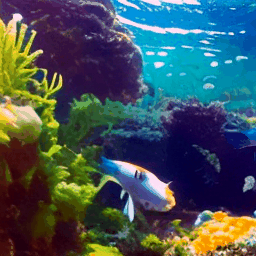}
        \includegraphics[width=0.115\linewidth]{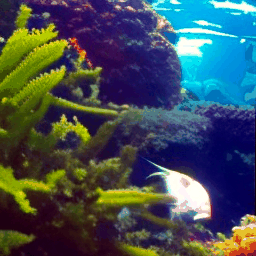}
        \includegraphics[width=0.115\linewidth]{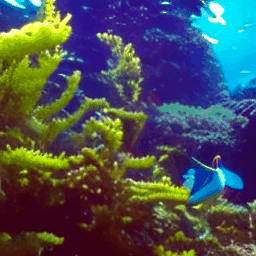}
        \includegraphics[width=0.115\linewidth]{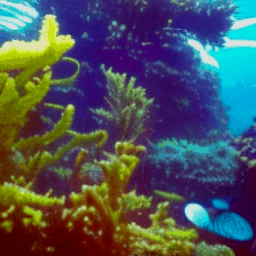}
        \includegraphics[width=0.115\linewidth]{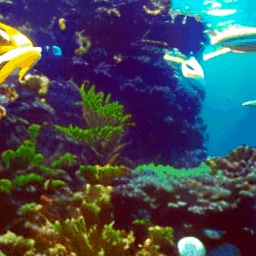}
    }
    \subcaptionbox{\textit{The camera gradually \textbf{zooms in} to focus on a running rabbit on the grassland. Then the camera \textbf{pans left} to give a panoramic shot, while the rabbit stops moving}.}
    {
        \includegraphics[width=0.115\linewidth]{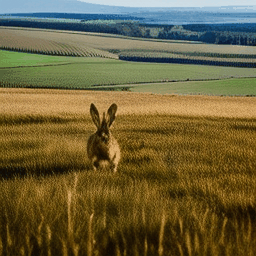}
        \includegraphics[width=0.115\linewidth]{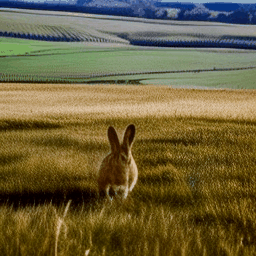}
        \includegraphics[width=0.115\linewidth]{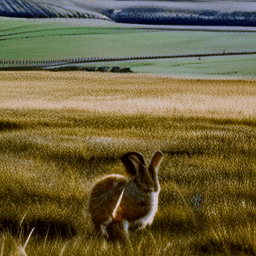}
        \includegraphics[width=0.115\linewidth]{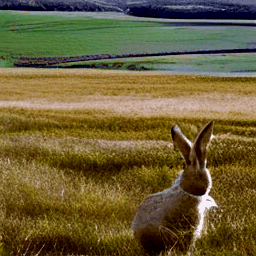}
        \includegraphics[width=0.115\linewidth]{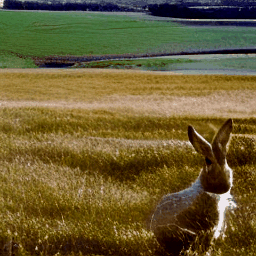}
        \includegraphics[width=0.115\linewidth]{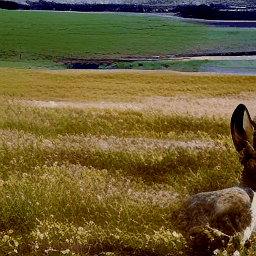}
        \includegraphics[width=0.115\linewidth]{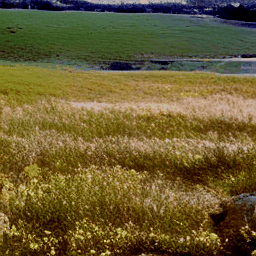}
        \includegraphics[width=0.115\linewidth]{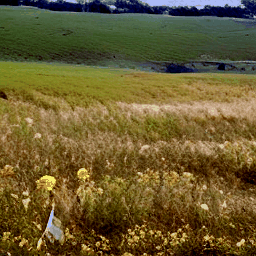}
        
    }
    \subcaptionbox{\textit{Starting with an indoor shot, the camera \textbf{moves left} to reveal a big window. Then the camera gradually \textbf{zooms in}, focusing on the trees outside}.}
    {
        \includegraphics[width=0.115\linewidth]{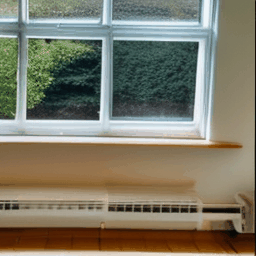}
        \includegraphics[width=0.115\linewidth]{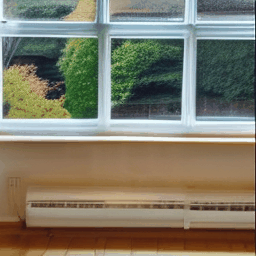}
        \includegraphics[width=0.115\linewidth]{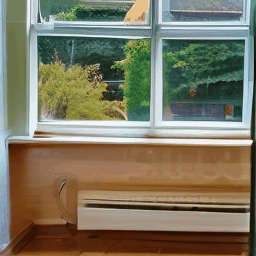}
        \includegraphics[width=0.115\linewidth]{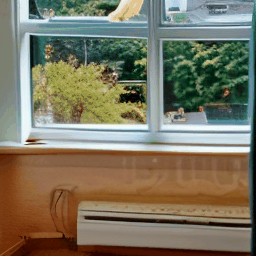}
        \includegraphics[width=0.115\linewidth]{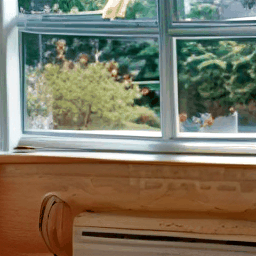}
        \includegraphics[width=0.115\linewidth]{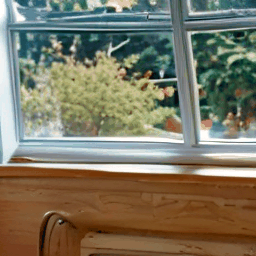}
        \includegraphics[width=0.115\linewidth]{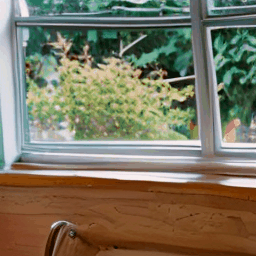}
        \includegraphics[width=0.115\linewidth]{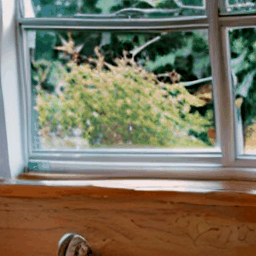}
    }
    \caption{Videos generated by Modular-Cam, where the user instruction is presented in the corresponding subcaption.}
    \label{fig:qualitative-result-supp}
\end{figure*}

\section{Training Procedure}
In this section, we give the detailed training procedure of Modular-Cam in Algorithm~\ref{alg:train}, which is a sequential optimization of the base video generator, CamOperator and AdaControlNet. 

Suppose we have training data in batches, that is $\mathbb{B}=\{\mathbb{B}_{VG},\mathbb{B}_{CO},\mathbb{B}_{AC}\}$, where $\mathbb{B}_{VG}$, $\mathbb{B}_{CO}$ and $\mathbb{B}_{AC}$ are training data for base video generator, CamOperator and AdaControlNet, respectively. $\mathbb{B}_{VG}$ includes video data $x^{1:f}\in\mathbb{R}^{b\times c\times f\times h\times w}$ together with its corresponding text description $y$, $\mathbb{B}_{CO}=\{\mathbb{B}_{CO}^{\textit{Motion}}\}$ are video data that follow specific $\textit{Motion}$ pattern, where $\textit{Motion}\in S_{\textit{Motion}}=\{\textit{ZoomIn},\textit{ZoomOut},\textit{PanLeft},\textit{PanRight},\textit{TiltUp},\textit{TiltDown}\}$, and $\mathbb{B}_{AC}$ further picks the starting frame of each video, i.e., $x_{cond}^{\textit{optional}}$, as the condition image. We denote the parameters for the pre-trained Stable Diffusion Model, base video generator, CamOperator and AdaControlNet as $\theta_{PD}$, $\theta_{VG}$, $\theta_{CO}$ and $\theta_{AC}$, respectively.

\begin{breakablealgorithm}
  \caption{Training of Modular-Cam}
  \label{alg:train}
  \begin{algorithmic}[1]
    \State \textbf{Input}: $\mathbb{B}=\{\mathbb{B}_{VG},\mathbb{B}_{CO},\mathbb{B}_{AC}\}$.
    \State \textbf{Parameters}: $\theta=\{\theta_{PD},\theta_{VG},\theta_{CO},\theta_{AC}\}$
    \Function{AddNoise}{$z_0^{1:f}, \epsilon^{1:f}, t$}
      \State $z_t^{1:f}\gets \sqrt{\alpha_t}z_0^{1:f}+\sqrt{1-\alpha_t}\epsilon^{1:f}$
      \State \Return $z_t^{1:f}$
    \EndFunction
    \Function{Optimize}{$\theta,\theta_{opt},\mathbb{B}$}{\small\Comment{$\theta_{opt}$ represents the parameters to be optimized}}
      \Repeat
        \For{$\{x^{1:f},y,x_{cond}^{\textsc{optional}}\}\in\mathbb{B}$}
          \State $z_0^{1:f}\gets \mathcal{E}(x^{1:f})$
          \State $\epsilon^{1:f}\sim\mathcal{N}(0,1)^{1:f}$
          \State \textsc{ts} $\gets$ \textsc{randint}{(0, \textsc{MAX\_TS}, \textsc{NUM\_TS)}}
          \For{$t$ in \textsc{ts}}
            \State $z_t^{1:f}\gets$\Call{AddNoise}{$z_0^{1:f},\epsilon^{1:f},t$}
            \If{$x_{cond}^{\textit{optional}}\neq$ \textsc{None}}
              \State $\epsilon_{pred}\gets \epsilon_{\theta}\left(z_t^{1:f},t,\tau_{\theta}(y),x_{cond}^{\textit{optional}}\right)$
            \Else
              \State $\epsilon_{pred}\gets \epsilon_{\theta}\left(z_t^{1:f},t,\tau_{\theta}(y)\right)$
            \EndIf
            \State $\mathcal{L}\gets\mathbb{E}_{z_t^{1:f},t,y}\left[||\epsilon-\epsilon_{pred}||_2^2\right]$
            \State $\theta_{opt}\gets\arg\min_{\theta_{opt}}\mathcal{L}$
          \EndFor
        \EndFor
      \Until{Converged}
      \State \Return $\theta_{opt}$
    \EndFunction
    \State \textbf{Begin Main Function}
    \State Initialize \textsc{MAX\_TS}, \textsc{NUM\_TS}
    \State $\theta\gets\{\theta_{PD}\}$
    \State $\theta_{VG}\gets$\Call{Optimize}{$\{\theta,\theta_{VG}\},\theta_{VG},\mathbb{B}_{VG}$}
    \State $\theta\gets\{\theta,\theta_{VG}\}$
    \For{$\textit{Motion}\in S_{\textit{Motion}}$}
      \State $\theta_{CO}^{\textit{Motion}}\gets$\Call{Optimize}{$\{\theta,\theta_{CO}^{\textit{Motion}}\},\theta_{CO}^{\textit{Motion}},\mathbb{B}_{CO}^{\textit{Motion}}$}
    \EndFor
    \State $\theta\gets\{\theta,\theta_{CO}\}$
    \State $\theta_{AC}\gets$\Call{Optimize}{$\{\theta,\theta_{AC}\},\theta_{AC},\mathbb{B}_{AC}$}
    \State $\theta\gets\{\theta,\theta_{AC}\}$
    \State \textbf{End Main Function}
  \end{algorithmic}
\end{breakablealgorithm}

\section{Additional Experimental Results}
In this section, we give additional results generated by Modular-Cam in Figure~\ref{fig:qualitative-result-supp} for further investigation. Specifically, we randomly generate some user instructions containing multiple scenes and guidance on camera-view transformations for generation. Note that we only display several intermediate frames for the convenience of presentation.

\section{Limitations and Future Work}
In this section, we discuss the possible limitations of our work. 
For the diversity of camera movements, on the one hand, these basic modules, e.g. ZoomIn, can be freely combined to constitute a more complicated pattern, e.g. ZoomIn and PanLeft. On the other hand, users can specify in the text to determine how far to PanRight, e.g. until the object out of view. The LLM can control the degree of variations by determining how long a clip needs to be generated.

However, for even finer controls, like exactly how much to PanRight or to rotation, we believe it can be solved by collecting data with finer conditioning information and data augmentation, simulating more diversified and complicated data. However, we note it may require more inputs, so it may be a trade off between user-friendliness and quality. We take it as future work, since we focus on creating dynamic camera-view for multi-scene videos.

\end{document}